\documentclass[conference,a4paper]{IEEEtran}
\usepackage{cite}
\usepackage{amsmath,amssymb,amsfonts}
\usepackage{graphicx}
\usepackage{textcomp}
\usepackage{xcolor}
\usepackage{float}
\usepackage{caption}
\usepackage{url}
\usepackage{hyperref}
\usepackage[T1]{fontenc}
\usepackage{cite}
\usepackage{amsmath,amssymb,amsfonts}
\usepackage{algorithm}
\usepackage{algpseudocode}
\usepackage{graphicx}
\usepackage{textcomp}
\usepackage{xcolor}
\usepackage{float}
\usepackage{caption}
\usepackage{url}
\usepackage{multirow}
\usepackage{hyperref}
\usepackage{svg}
\usepackage[symbol]{footmisc}

\def\BibTeX{{\rm B\kern-.05em{\sc i\kern-.025em b}\kern-.08em
    T\kern-.1667em\lower.7ex\hbox{E}\kern-.125emX}}
\begin{document}

\title{A Computational Approach to Modeling Conversational Systems: Analyzing Large-Scale Quasi-Patterned Dialogue Flows}
\author{
    \IEEEauthorblockN{1\textsuperscript{st} Mohamed Achref Ben Ammar}
    \IEEEauthorblockA{
        \textit{Software Engineering and Mathematics Department} \\
        \textit{National Institute of Applied Science and Technology} \\
        Tunis, Tunisia \\
        mohamedachref.benammar@insat.ucar.tn
    }
    \and
    \IEEEauthorblockN{2\textsuperscript{nd} Mohamed Taha Bennani}
    \IEEEauthorblockA{
        \textit{Computer Science Department}\\
        \textit{University of Tunis El Manar} \\
        Tunis, Tunisia \\
        taha.bennani@fst.utm.tn
    }
}


\maketitle

\begin{abstract}
The analysis of conversational dynamics has gained increasing importance with the rise of large language model-based systems, which interact with users across diverse contexts. In this work, we propose a novel computational framework for constructing conversational graphs that capture the flow and structure of loosely organized dialogues, referred to as quasi-patterned conversations. We introduce the Filter \& Reconnect method, a novel graph simplification technique that minimizes noise while preserving semantic coherence and structural integrity of conversational graphs. Through comparative analysis, we demonstrate that the use of large language models combined with our graph simplification technique has resulted in semantic metric S increasing by a factor of 2.06 compared to previous approaches while simultaneously enforcing a tree-like structure with 0 \(\delta\)-hyperbolicity, ensuring optimal clarity in conversation modeling. This work provides a computational method for analyzing large-scale dialogue datasets, with practical applications related to monitoring automated systems such as chatbots, dialogue management tools, and user behavior analytics.
\end{abstract}


\section{Introduction}

With the rise of large language model \cite{zhao2024surveylargelanguagemodels , radford2018improving, vaswani2017attention} (LLM)-based conversational systems, the analysis of conversational dynamics has gained increasing importance. Understanding how conversations flow, identifying underlying intents, and modeling dialogue structures are critical for improving the effectiveness of these systems in real-world applications. Recent advancements in natural language processing (NLP) \cite{carrasco2024nlp} have facilitated progress in conversational systems modeling techniques such as intent extraction \cite{schuster-etal-2019-cross-lingual} and dialogue act modeling \cite{khanpour-etal-2016-dialogue}. While dialogue act modeling classifies utterances into communicative functions using models like Long Short-Term Memory (LSTM) networks \cite{hochreiter1997long}, intent extraction identifies the underlying intent of an utterance through models like biLSTM-CRF \cite{huang2015bidirectional}. However, these techniques primarily focus on analyzing conversations in isolation rather than exploring recurring patterns across sets of conversations.

Recent studies have introduced methods for constructing conversational graphs to analyze dialogue dynamics. Ferreira et al. (2023) \cite{ferreira2023unsupervised} presented an unsupervised flow discovery method for task-oriented dialogues, utilizing frequent verb phrases extracted via the spaCy toolkit \cite{spacy2} and keywords identified through KeyBERT to represent dialogue states. This approach constructs transition graphs to model dialogue flows without relying on LLMs or sentiment analysis.

In this work, we propose a computational approach for constructing conversational graphs, where each node represents an intent and directed edges indicate transitions labeled with their respective probabilities.

Our approach focuses on quasi-patterned sets of conversations, which are collections of dialogues that, while not adhering to strict conversational structures, still display common underlying patterns in the flow of intents. These conversations may vary in specifics, but they tend to follow similar intent transitions and themes, reflecting consistent conversational dynamics across multiple interactions.

The primary steps of our approach include utterance embedding using Sentence-BERT (SBERT) \cite{DBLP:journals/corr/abs-1908-10084}, Kmeans++ clustering \cite{arthur2007kmeans}, and intent extraction through LLMs \cite{rodriguez2023intentgpt , song2023openworld , he2023intent , mirza2024illuminer , kim2024autointent}. Following these steps, we construct a transition matrix and generate the conversational graph. These processes will be thoroughly detailed in the following sections of this paper.

This paper is organized as follows: A review of relevant work is given in Section II. The method is then described in Section III, followed by a case study of our approach in Section IV and the assessment metrics in Section V. The findings and discussion are presented in Section VI, and the study's conclusions are provided in Section VII.

\section{Related Works}

The study of conversational systems has evolved from early rule-based systems \cite{rulebasedintentextraction} to modern deep learning approaches for intent extraction and dialogue act modeling \cite{schuster-etal-2019-cross-lingual, devlin-etal-2019-bert}. Despite improvements, scaling these models to diverse interactions remains challenging.

\subsection{Intent Extraction and Dialogue Act Modeling}

Intent extraction identifies the goal behind a user’s utterance. Early methods such as biLSTM-CRF \cite{huang2015bidirectional} and RNNs \cite{hochreiter1997long} have been largely replaced by Transformer-based models like BERT \cite{devlin-etal-2019-bert} and GPT-based models \cite{radford2018improving,zhao2024surveylargelanguagemodels}, which offer better contextual understanding but require large annotated datasets and struggle with domain adaptation \cite{song2023openworld}. Dialogue act modeling, which classifies utterances (e.g., questions, statements), also transitioned from statistical methods to deep learning techniques \cite{khanpour-etal-2016-dialogue, hochreiter1997long, schuster-etal-2019-cross-lingual}. However, both approaches typically analyze utterances in isolation, missing recurring conversational patterns.

\subsection{Graph-Based Dialogue Analysis}

Graph-based methods analyze dialogue flows by constructing transition graphs from clustered utterances. Ferreira et al. \cite{ferreira2023unsupervised} proposed an unsupervised framework using spaCy \cite{spacy2} for frequent verb phrase extraction and KeyBERT \cite{grootendorst2020keybert} for keyword identification. Their method suffers from vague cluster labels and noisy graphs due to the absence of a structured simplification mechanism.

\subsection{Our Contributions}

Our approach improves conversational modeling by:
\begin{itemize}
    \item \textbf{Enhanced Intent Extraction:} Leveraging LLMs \cite{rodriguez2023intentgpt, he2023intent, mirza2024illuminer, kim2024autointent} to generate more coherent and semantically rich intent labels than verb-phrase-based methods \cite{ferreira2023unsupervised}.
    \item \textbf{Graph Simplification via Filter \& Reconnect:} Introducing a novel technique that removes redundant transitions while preserving essential conversational structures, resulting in cleaner, more interpretable dialogue graphs.
\end{itemize}

By combining LLM-based intent extraction with structured graph simplification, our method significantly enhances the analysis of large-scale dialogue datasets, providing clearer insights into conversational patterns while improving practical applications such as chatbot monitoring and user behavior analysis.

\section{Methodology}

In this section, we present the methodology of our approach. The process begins by extracting utterances from the dataset, then each utterance is embedded using a text-embedding model \cite{DBLP:journals/corr/abs-1908-10084}, thus transforming them into vectors that represent the semantic meaning of each utterance.

These vectors are then clustered to identify the key utterance intents contained in the quasi-patterned set of conversations using an LLM. After that, we construct a Markov Chain, which will be filtered and processed to discard the noise and irrelevant intent transitions from the graph, and thus we create a conversational graph that can be analyzed to extract conversational flow patterns.

\subsection{Vector Embedding Generation}
The first step involves generating vector embeddings for each utterance using a pre-trained text embedding model\cite{DBLP:journals/corr/abs-1908-10084}. Formally, let \(U=\{u_1, u_2, \dots, u_N\}\) be the set of utterances. A pre-trained embedding model \( f: U \to \mathbb{R}^d \) maps each utterance \( u_i \) to a vector \(\mathbf{x}_i = f(u_i)\). By doing so, we capture semantic and contextual properties of the utterances in a continuous metric space. For this work, we utilized the all-MiniLM-L12-v2 model \cite{DBLP:journals/corr/abs-1908-10084}, which converts textual data into high-dimensional vectors \(\mathbf{x}_i \in \mathbb{R}^d\) that represent the semantic meaning of the utterances. These embeddings form the foundation for clustering, as they encapsulate both syntactic and semantic similarities \cite{padmasundari2018intent}. By using embeddings that capture the contextual meaning of utterances, we ensure that the clusters formed represent coherent thematic groups.

\subsection{Vector Clustering }
Once the vector embeddings \(\{\mathbf{x}_i\}_{i=1}^N\) of the utterances are generated, we use K-means++ \cite{arthur2007kmeans} for clustering. Let \(N_c\) be the number of clusters. K-means aims to find \(\{\boldsymbol{\mu}_c\}_{c=1}^{N_c}\) that minimize:
\begin{equation}
    \min_{\{\boldsymbol{\mu}_c\}} \sum_{i=1}^{N} \min_{1 \leq c \leq N_c} \|\mathbf{x}_i - \boldsymbol{\mu}_c\|^2 
\end{equation}

K-means++ strategically initializes centroids \(\boldsymbol{\mu}_c \in \mathbb{R}^d\) to be well-spread, leading to faster convergence and more accurate clusters. This reduces the likelihood of poor clustering due to random initialization, while maintaining the simplicity and scalability of K-means. The algorithm is efficient for large datasets, making it ideal for grouping utterances into meaningful thematic categories with minimal computational cost.

\subsection{Intent Extraction}
After clustering, we proceed with intent extraction from the clusters. To achieve this, we select the \( n \) vectors, closest to the centroid of each cluster and extract the common intent from the corresponding utterances. 
For intent extraction, we leverage the gemini-1.5-flash model \cite{geminiteam2024gemini15unlockingmultimodal} to summarize the core intent of each cluster, enabling us to label the groups with meaningful thematic descriptions.

\subsection{Markov Chain Construction}
Following clustering and intent extraction, we build a Markov chain \cite{norris1998markov} to model the transition probabilities between clusters based on the conversations' flow. A Markov chain is a stochastic model that describes transitions from one state to another according to defined probabilities. In this context, the states correspond to the different clusters, and the transitions between them represent the flow of intents throughout the conversations.

Let \( S = \{s_1, s_2, \dots, s_N\} \) denote the sequence of cluster IDs assigned to the \( N \) utterances in the order they appear in the conversations. We construct a transition matrix \( T \in \mathbb{R}^{N_c \times N_c} \), where \( N_c \) is the number of unique clusters. The matrix element \( T_{i,j} \) is defined as:
\begin{equation}
    T_{i,j} = \frac{\text{count}(s_{t} = i, s_{t+1} = j)}{\sum_{j=1}^{N_c} \text{count}(s_{t} = i, s_{t+1} = j)}
\end{equation}

where \(\text{count}(s_{t}=i, s_{t+1}=j)\) is the frequency of transitions from cluster \( i \) to cluster \( j \) observed in the dataset. Therefore, \( T_{i,j} \) expresses the probability of moving from cluster \( i \) to cluster \( j \).

The transition matrix \( T \) encapsulates the probabilistic flow of intents, representing the natural progression of themes and topics across the dataset. This Markov chain model maps how conversational elements evolve, providing insights into the underlying recurring patterns of the conversations; it will be considered as our base conversational graph for the next steps.

\subsection{Conversational Graph Processing}
After the Markov chain is constructed, we refine the conversational graph using three different processing techniques: filter \& reconnect, top-k filtering, and threshold filtering. Each technique builds on the one before it. This stage guarantees the removal of noise edges and the preservation of the quasi-patterned set of conversations' structure. We represent the conversational graph as \(G = (V, E)\), where \( V = \{1, \dots, N_c\} \) is the set of cluster nodes and \( E \) consists of directed edges weighted by the transition probabilities \( T_{i,j} \). Each technique simplifies the graph in different ways, aiming to improve readability and interpretability while maintaining meaningful conversational paths.

\subsubsection{Threshold Technique} 
This method simplifies the graph by removing edges with weights below a certain threshold \(\tau\). 
A higher threshold \(\tau\) results in a sparser graph with fewer, more significant transitions, while a lower \(\tau\) retains more transitions but may introduce noise.

\subsubsection{Top-K Filtering Technique}
Building on the threshold technique, the Top-K filtering method not only enforces a minimum weight threshold \(\tau\) but also restricts the number of edges connected to each node, giving us direct control over the degree of the graph. For each node \( i \), only the \(K\) highest-weighted edges are retained, ensuring that each node is connected only to its most significant neighbors.
This approach not only filters out low-confidence transitions but also explicitly limits the degree of each node, resulting in a sparser and more manageable graph that focuses on the most likely transitions.

\subsubsection{Filter \& Reconnect Method}
Building on the foundational techniques of Threshold filtering and Top-\(K\) filtering, the Filter \& Reconnect method refines the conversational graph by ensuring an acyclic structure while maintaining interpretability. The method proceeds in three stages:

\begin{itemize}

    \item \textbf{Filtering:}  
    First, we remove edges with weights below a threshold \(\tau\) as well as self-transitions. In addition, for each node \(i\), we retain only the top \(K\) strongest incoming edges. 
    
    \item \textbf{Cycle Removal:}  
        Next, we identify any cycles present in \(G_{K,\tau}\). For each cycle \(C\), we remove the weakest edge—i.e., the edge \((i,j) \in C\) with the smallest weight \(T_{i,j}\)—to break the cycle. This process is applied iteratively until the graph becomes acyclic.

    \item \textbf{Reconnection:}  
        During the previous steps, some subgraphs may become disconnected from the main graph. To preserve a coherent, tree-like structure, we reconnect these subgraphs by restoring the strongest transitions that were previously removed.

\end{itemize}

By combining Threshold filtering and Top-\(K\) filtering, this method provides precise control over the graph's degree and sparsity. The ensuing cycle removal and reconnection procedures guarantee that the finished conversational graph is linked and retains a tree-like structure, as demonstrated in the following Case Study section.
\section{Case Study}

\begin{figure}[h]
\centering
\captionsetup{justification=centering} 
\includegraphics[width=1\linewidth]{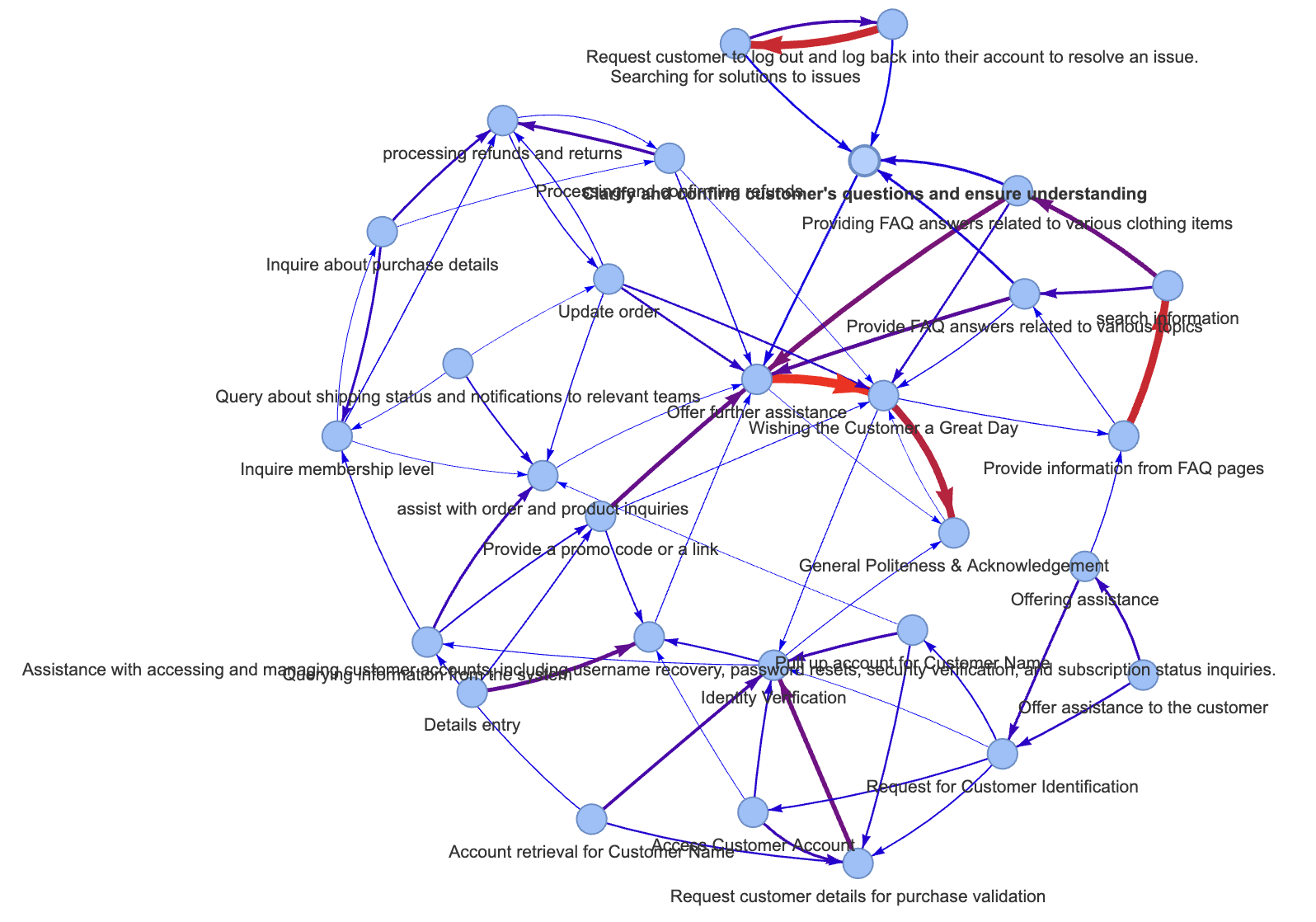}
\caption{Conversational graphs generated using threshold filtering for \(\tau\) = 0.1 on the ABCD dataset}
\label{fig:threshold_graphs}
\end{figure}

\begin{figure}[h]
\captionsetup{justification=centering} 
\centering
\includegraphics[width=0.89\linewidth]{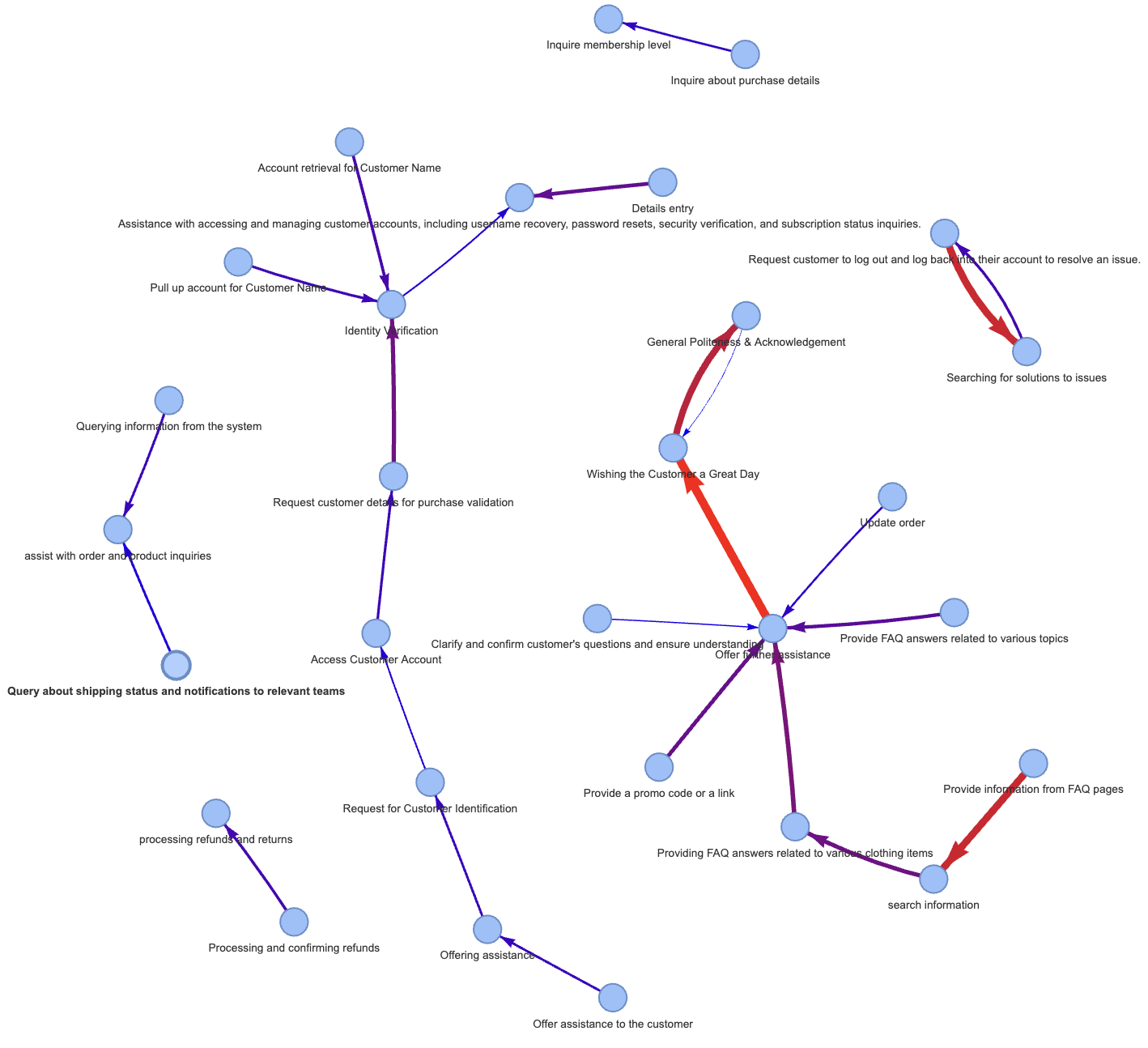}
\caption{Conversational graphs generated using Top-K filtering for top-k = 1 and \(\tau\) = 0.1 on the ABCD dataset}
\label{fig:top_k_graphs}
\end{figure}

\begin{figure}[h!]
\centering
\captionsetup{justification=centering}
\includegraphics[width=1\linewidth]{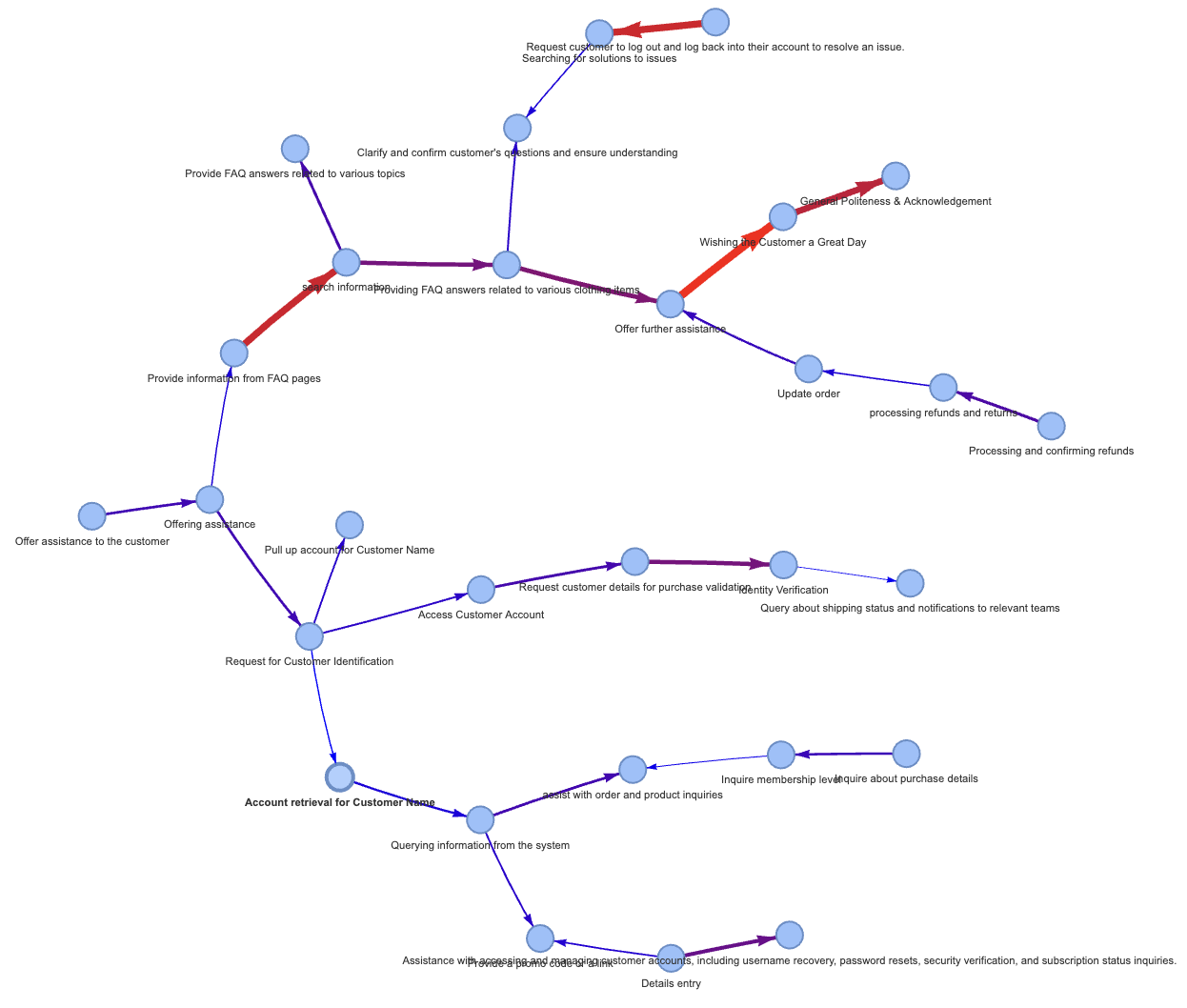}
\caption{Conversational graph generated using the Filter \& Reconnect method for top-k = 1 and \(\tau\) = 0.1 on the ABCD dataset}
\label{fig:filter_and_reconnect_graph}
\end{figure}

This section demonstrates our approach on the Action-Based Conversations Dataset (ABCD) \cite{chen2021abcd}, visualizing the progressive refinement of conversational graphs through Threshold Filtering, Top-K Filtering, and our novel Filter \& Reconnect method.  Threshold Filtering, the initial step, reduces noise by eliminating low-weight edges, resulting in a less noisy graph as shown in Figure \ref{fig:threshold_graphs}.
Building upon this, Top-K Filtering refines the graph further by controlling the node degree.  It retains only the most significant connections for each node, creating a cleaner graph but potentially fragmenting it into disconnected subgraphs and doesn't address the problem of cycles as shown in Figure \ref{fig:top_k_graphs}.
Finally, Filter \& Reconnect builds on the logic of both previous steps.  It takes the cleaner, degree-controlled graph from Top-K Filtering and addresses its limitations.  First, it systematically breaks cycles to create an acyclic structure. Then, it strategically reconnects previously fragmented subgraphs by restoring the strongest removed edges. This crucial reconnection step ensures that the final graph maintains a coherent, tree-like structure, capturing the dominant intent transitions within the ABCD dataset while enhancing interpretability.  The resulting graph, shown in Figure \ref{fig:filter_and_reconnect_graph} highlights the capabilities of our Filter \& Reconnect approach for analyzing loosely structured dialogues and representing them in a clear, tree-like format.

\section{Evaluation Metrics}
We evaluate our approach through two categories of metrics: structural metrics and semantic similarity metrics. These metrics provide a comprehensive analysis of the readability, interpretability, and representational quality of the conversational graphs.

\subsection{Structural Metrics}
Structural metrics evaluate the topological and structural characteristics of the graph, with a focus on its readability and alignment with the quasi-patterned nature of conversations. The metrics considered are defined as follows:

\subsubsection{Branching Factor (\(B\))}
The branching factor is defined as the average number of edges per vertex in the graph. Let \(G = (V, E)\) be the graph, where \(V\) is the set of vertices and \(E\) is the set of edges.
A higher branching factor (\(B\)) may indicate more complex dialogue paths, while a lower value suggests a simpler, more linear structure.

\subsubsection{Number of Cycles (\(C\))}
The number of cycles \cite{onnumberofcycles} in the graph is a direct measure of its cyclicity, which affects its interpretability. A cycle in the graph is defined as a closed path with no repeated vertices or edges, except for the starting and ending vertices.
Graphs with fewer cycles are more readable, as they resemble the linear and branching structures typical of conversational flows.
    
\subsubsection{\(\delta\)-Hyperbolicity}
\(\delta\)-Hyperbolicity \cite{gromovhyperbolic , doi:10.1080/15427951.2013.828336} measures how close the graph is to being Gromov-hyperbolic, which quantifies its tree-like properties. Formally, for any four vertices \(u, v, w, x \in V\), consider the metric space induced by the shortest-path distances \(d(u, v)\).  Let 
    \begin{align}
        S_1 &= d(u,v) + d(w,x), \\
        S_2 &= d(u,w) + d(v,x), \\
        S_3 &= d(u,x) + d(v,w).
    \end{align}

    Let \(M_1\) and \(M_2\) be the two largest values among \(S_1\), \(S_2\), and \(S_3\).  Define \(\text{hyp}(u, v, w, x) = M_1 - M_2\). The \(\delta\)-hyperbolicity is defined as:
    \begin{equation}
        \delta(G) = \frac{1}{2} \max_{u, v, w, x \in V} \text{hyp}(u, v, w, x)    
    \end{equation}
where the maximum is taken over all possible sets of four vertices. Lower \(\delta\) values indicate a structure closer to a tree, aligning with the natural branching observed in conversational graphs.

\subsection{Semantic Metrics}
Semantic similarity metrics evaluate how well the constructed conversational graph captures the underlying dialogue flow and semantics of the dataset. These include:

\subsubsection{Cosine Similarity of Utterance Assignments}
    This metric measures the semantic alignment between each cluster's centroid (representing the cluster's intent) and the embeddings of the utterances assigned to that cluster. Define a function
    \begin{equation}
        \mu(c) : c \in \mathcal{C} \rightarrow \mathbb{R}^d
    \end{equation}
    that returns the centroid embedding for cluster \(c\). For each utterance \(u_i\) with embedding \(\mathbf{x}_i\) assigned to cluster \(c\), the cosine similarity is defined as:
    \begin{equation}
        \text{Similarity}(u_i, c) = \frac{\mathbf{x}_i \cdot \mu(c)}{\|\mathbf{x}_i\|\,\|\mu(c)\|}
    \end{equation}
    The overall semantic coherence is calculated as the average similarity over all utterances:
    \begin{equation}
    S = \frac{1}{N} \sum_{c \in \mathcal{C}} \sum_{u_i \in I_c} \frac{\mathbf{x}_i \cdot \mu(c)}{\|\mathbf{x}_i\|\,\|\mu(c)\|}  
    \end{equation}
    
    where \(I_c\) is the set of utterances in cluster \(c\) and \(N\) is the total number of utterances. A higher \(S\) indicates that the utterances within a cluster are well-aligned with the cluster centroid, reflecting the quality of the clustering and intent extraction processes.

\subsubsection{Coverage}
    This metric quantifies the effectiveness of the conversational graph \(G = (V, E)\) in representing the actual transitions observed in the identified intent flows in the dataset \(F\). The graph vertices \(V\) represent the extracted intents, and the edges \(E\) represent the transition probabilities between them. 
    
    For a conversation flow \(f = [i_1, i_2, \dots, i_n]\), the \textit{Transition Alignment Score} (TAS) is defined as:
    \begin{equation}    
        \text{TAS}(f) = \frac{\sum_{j=1}^{n-1} \mathbb{1}((i_j, i_{j+1}) \in E)}{n-1} 
    \end{equation}
    where \(\mathbb{1}((i_j, i_{j+1}) \in E)\) is an indicator function equal to 1 if the transition \((i_j, i_{j+1})\) exists in the graph \(G\), and 0 otherwise.
    
    The overall \textit{Coverage Metric} \(C\) is then defined as the average TAS across all flows \(f \in F\):
    \begin{equation}
     C = \frac{1}{|F|} \sum_{f \in F} \text{TAS}(f)       
    \end{equation}
    A higher coverage value indicates that the graph accurately captures the real-world conversational flows, representing a larger proportion of observed transitions.

By combining these semantic similarity metrics, we assess how well the conversational graph captures the flow and thematic coherence of the quasi-patterned dialogue dataset.

\section{Results and Discussion}

In this section, we present a comprehensive benchmarking of conversational modeling techniques, including our proposed methods and the baseline approach by Ferreira et al. (2023) \cite{ferreira2023unsupervised}, across multiple conversational datasets.

\subsection{Datasets}
To ensure a comprehensive evaluation, we employed two datasets that represent different conversational dynamics:

\subsubsection{MultiWOZ 2.2 Dataset}
This dataset \cite{zang2020multiwoz} is a large-scale, multi-domain task-oriented dialogue dataset consisting of structured, goal-driven conversations. The dataset includes dialogues across domains such as hotel bookings, restaurant reservations, and taxi services. Its well-annotated nature and clear intent transitions make it suitable for evaluating structured conversational graph construction methods.

\subsubsection{ABCD Dataset}
The Action-Based Conversations Dataset (ABCD) \cite{chen2021abcd} contains human-to-human conversations designed for action-oriented tasks, such as troubleshooting and customer service. Unlike MultiWOZ 2.2, ABCD features loosely structured dialogues with quasi-patterned flows.

\subsection{Performance Analysis}

Table \ref{tab:results} presents a comparative analysis of different conversational graph construction methods across two datasets: MultiWOZ 2.2 \cite{zang2020multiwoz} and  ABCD \cite{chen2021abcd}.  The table evaluates four approaches: Ferreira et al. \cite{ferreira2023unsupervised}, Threshold Filtering, Top-k Filtering, and Filter \& reconnect. The table is organized by dataset, with results for each approach displayed in separate rows.  Boldface indicates the best performance for each metric within each dataset. The key findings are discussed below:

\begin{table}[h!]
\centering
\renewcommand{\arraystretch}{1.2} 
\setlength{\tabcolsep}{6pt}      
\begin{tabular}{||l||c|c|c|c|c||}
\hline\hline
 & \multicolumn{5}{c||}{\textbf{MultiWOZ 2.2 dataset}} \\
\cline{2-6}

\textbf{Approach}  & 
\multicolumn{3}{c|}{\textbf{structural metrics}} &  
\multicolumn{2}{c||}{\textbf{semantic metrics}}  \\
\cline{2-6}

 & \textbf{B} 
 & \textbf{\(N_\text{cycles}\)} 
 & \boldmath$\delta$
 & \textbf{S}  
 & \textbf{C} 
\\
\hline\hline

\textbf{Ferreira et al. (2023)}     
 & 3.8 & 14 & 1.5 & 0.122 & \textbf{1.00}\\
\hline
\hline
\textbf{Threshold Filtering}       
 & 2.71 & 13 & 1.5 & 0.373 & 0.497\\
\hline
\textbf{Top\_k Filtering}           
 & 0.95 &  1 & 0.02  & 0.379 &  0.161\\
\hline

\textbf{Filter \& reconnect}      
 & \textbf{0.89} & \textbf{0}  &  \textbf{0} & \textbf{0.384} & 0.175  \\
\hline\hline
 & \multicolumn{5}{c||}{\textbf{ABCD dataset}} \\
\cline{2-6}

\textbf{Approach}  & 
\multicolumn{3}{c|}{\textbf{structural metrics}} &  
\multicolumn{2}{c||}{\textbf{semantic metrics}}  \\
\cline{2-6}

 & \textbf{B} 
 & \textbf{\(N_\text{cycles}\)} 
 & \boldmath$\delta$
& \textbf{S}  
 & \textbf{C} 

\\
\hline\hline

\textbf{Ferreira et al. (2023)}     
 & 3 & 26 & 1.5 & 0.148  & \textbf{1.00} \\
\hline
\hline
\textbf{Threshold Filtering}       
 &  2.67 & 16 & 1.5 &  0.295 & 0.522 \\
\hline
\textbf{Top\_k Filtering}           
 & 0.95 & 2 &  0.01 & 0.288 & 0.156\\
\hline
\textbf{Filter \& reconnect}      
 & \textbf{0.78} & \textbf{0} & \textbf{0}  & \textbf{0.305} &  0.172\\
\hline

\end{tabular}
\captionsetup{justification=centering}
\caption{Benchmarking of approaches on the MultiWOZ 2.2 and ABCD datasets}
\label{tab:results}
\end{table}

\subsubsection{Structural Metrics}
Ferreira et al.'s method retains all transitions, resulting in high complexity—e.g., on the ABCD dataset, a branching factor (B) of 3, 26 cycles, and \(\delta\)-Hyperbolicity of 1.5—making the graph hard to interpret. Threshold Filtering reduces complexity (B = 2.67, 16 cycles on ABCD) but still yields a high \(\delta\). Top-K Filtering further lowers complexity (B = 0.95, 2 cycles, \(\delta\) = 0.01) yet produces disconnected subgraphs. In contrast, Filter \& Reconnect delivers the simplest, tree-like structure (B = 0.78, 0 cycles, \(\delta\) = 0) that is highly interpretable.

\subsubsection{Semantic Metrics}
Ferreira et al.'s approach achieves full coverage (C = 1.0) but with high noise. While Threshold and Top-K Filtering reduce coverage (e.g., 0.497 on MultiWOZ 2.2), our LLM-based intent extraction maintains consistently high semantic similarity. On the ABCD dataset, Filter \& Reconnect achieves an S value of 0.305 versus 0.148 by Ferreira et al. Note that filtering affects only edge selection, leaving node labels—and thus semantic similarity—largely unchanged.

\subsubsection{Discussion}
Overall, these results reveal trade-off between complete coverage and graph interpretability. Ferreira et al.’s method captures every transition but suffers from a higher branching factor and higher \(\delta\)-Hyperbolicity leading to a less interpretable and a more complex conversational graph. Also, it performs badly in semantic coverage due to its basic most common verb phrase cluster labeling. On the other hand, Filter \& Reconnect, powered by LLM-based intent extraction and its filtering method, offers a more balanced and semantically refined view of conversational flows, making it a strong candidate for analyzing complex or loosely structured dialogue datasets.

\section{Conclusion}
In this study, we presented a novel computational approach to modeling quasi-patterned conversational flows using conversational graphs. By applying advanced text embedding techniques and clustering methods, we effectively extracted conversational intents and transitions to build interpretable graph representations. Through the comparative analysis of multiple graph simplification methods—Threshold Filtering, Top-K Filtering, and Filter \& Reconnect—we demonstrated that the Filter \& Reconnect method yielded the most readable and insightful graphs. This method preserves meaningful transitions between conversational intents while eliminating noise, resulting in a tree-like structure that aligns well with the quasi-patterned nature of the dataset.

Our approach offers a solution for analyzing large-scale dialogue datasets, providing valuable insights into conversational dynamics that can be leveraged to enhance automated systems, such as customer support LLM-powered agents and dialogue management systems. The successful application of our method to the ABCD and MultiWOZ 2.2 datasets highlights its potential to uncover underlying conversational structures in other large-scale, loosely structured datasets, opening the door to further developments in conversational system optimization. The code for this paper is available on GitHub.
\footnote[2]{Code Repository : https://github.com/achrefbenammar404/quasi-patterned-conversations-analysis.git}

\end{document}